%% file: Interspeech_2019.tex
\documentclass[a4paper]{article}

\usepackage{INTERSPEECH2019}
\usepackage{amsmath,graphicx}
\usepackage{dsfont}
\usepackage{subcaption}
\usepackage{multirow}
\usepackage{hyperref}

\title{Subword RNNLM Approximations for Out-Of-Vocabulary Keyword Search}
\name{Mittul Singh$^*$, Sami Virpioja$^\dagger$, Peter Smit$^{*\ddagger}$, Mikko Kurimo$^*$\thanks{This work was supported by the Academy of Finland (Flagship programme: Finnish Center for Artificial Intelligence, FCAI; Grants 320181, 320182, 320183).}}
\address{$^*$Department of Signal Processing and Acoustics, Aalto University, Espoo, Finland\\
        $^\dagger$Department of Digital Humanities, Helsinki University, Helsinki, Finland\\
        $^\ddagger$Inscripta, Helsinki, Finland}
\email{firstname.lastname@\{aalto,helsinki\}.fi}

\begin{document}

\maketitle

\begin{abstract}
\input{abstract.tex}
\end{abstract}
\noindent\textbf{Index Terms}: OOV, Keyword Search, Single character, RNNLM, first-pass

\input{introduction.tex}

\input{related_work.tex}

\input{variprob.tex}

\input{experimental_setup.tex}
\input{kws_exp.tex}

\input{conclusion.tex}

\bibliographystyle{IEEEtran}

\bibliography{mybib}
\end{document}

%% file: abstract.tex
In spoken Keyword Search, the query may contain out-of-vocabulary (OOV) words not observed when training the speech recognition system. Using subword language models (LMs) in the first-pass recognition makes it possible to recognize the OOV words, but even the subword \textit{n}-gram LMs suffer from data sparsity. Recurrent Neural Network (RNN) LMs alleviate the sparsity problems but are not suitable for first-pass recognition as such. One way to solve this is to approximate the RNNLMs by back-off \textit{n}-gram models. In this paper, we propose to interpolate the conventional \textit{n}-gram models and the RNNLM approximation for better OOV recognition. Furthermore, we develop a new RNNLM approximation method suitable for subword units: It produces variable-order \textit{n}-grams to include long-span approximations and considers also \textit{n}-grams that were not originally observed in the training corpus. To evaluate these models on OOVs, we setup Arabic and Finnish Keyword Search tasks concentrating only on OOV words. On these tasks, interpolating the baseline RNNLM approximation and a conventional LM outperforms the conventional LM in terms of the Maximum Term Weighted Value for single-character subwords. Moreover, replacing the baseline approximation with the proposed method achieves the best performance on both multi- and single-character subwords.

%% file: introduction.tex
\section{Introduction}
The goal of Keyword Search (KWS) on audio data is to search for interesting terms (words or their sequences) in speech. These systems typically use an Automatic Speech Recognition (ASR) system in the background. The ASR system always fails to recognize words missing from its training data and thus finding interesting keywords containing these Out-of-Vocabulary (OOVs) words is a difficult task.\\
\indent Projects funded to develop and improve Keyword Search, like the IARPA's BABEL program, have promoted a lot of work to improve the OOV Keyword Search \cite{DBLP:conf/interspeech/LoganT02,6707766,DBLP:conf/icassp/ManguKSKP14,8536457,4777893,DBLP:conf/interspeech/HartmannLMLG14,bulyko2012subword,karakos2014subword,DBLP:journals/taslp/HeBFHJOFP16,DBLP:conf/interspeech/YuS04,DBLP:conf/icassp/SeideYMC04,DBLP:conf/interspeech/LeeT016}. These studies have improved KWS by handling OOV keywords in mostly two ways. First, by replacing OOVs by their acoustically-similar proxies \cite{DBLP:conf/interspeech/LoganT02,6707766,DBLP:conf/icassp/ManguKSKP14,8536457} and second, by employing subword units instead of words to be able to recognize OOVs \cite{4777893,DBLP:conf/interspeech/HartmannLMLG14,bulyko2012subword,karakos2014subword,DBLP:journals/taslp/HeBFHJOFP16,DBLP:conf/interspeech/YuS04,DBLP:conf/icassp/SeideYMC04,DBLP:conf/interspeech/LeeT016,Khokhlov2017}. In the latter approach, subword-based conventional \textit{n}-gram language models (LMs), with context sizes 2 to 4, have been successfully applied to first-pass recognition for OOV KWS. Here, using subword units is crucial to model OOVs as with word-based LMs OOVs are lost and cannot be recovered in subsequent recognition passes.\\
\indent Even with subword units, the conventional \textit{n}-gram LM has limitations. They suffer from data sparsity issues leading to inaccurate scoring and hence, assigning low or failing to detect good hypotheses in the first or subsequent recognition passes \cite{Deoras2011}. In effect, causing KWS to not recognize OOV keywords. Using long-span neural network LMs, such as Recurrent Neural Networks (RNNs), can help with data sparsity, but they are prohibitively expensive to use in first-pass decoding \cite{Deoras2011}. Thus, researchers apply these models in the first-pass by approximating them to \textit{n}-gram LMs \cite{Deoras2011,Adel2014,Arisoy:2014:CNN:2583704.2583722,8683481}. Inspired by these efforts, we develop a new method for RNNLM approximation to \textit{n}-gram LMs for first-pass decoding in subword KWS.\\
\begin{table}[t]
  \footnotesize
  \centering
  \caption{\label{tab:oov_lengths} The table displays the length rates in \#subwords per occurrence for frequent words (FW, training set frequency $\ge 5$) on the training and test sets, and for OOV words on the test set from the Arabic and Finnish KWS described in Section \ref{sec:setup}. OOV Rate on these datasets is 2\% approx. Length rates for both single- and multi-character subwords are presented.}
  \vspace{-0.3cm}
  \begin{tabular}{l c c c c c c}
    \hline
    \multirow{2}{*}{Language} & \multicolumn{2}{c}{FW (Train)} & \multicolumn{2}{c}{FW (Test)}& \multicolumn{2}{c}{OOV (Test)}\\
    & Single & Multi & Single & Multi & Single & Multi\\
    \hline
    Arabic   & 4.67 & 1.63 & 4.49 & 1.49 & \textbf{6.80} & \textbf{3.32} \\
    Finnish  & 7.03 & 1.08 & 7.25 & 1.07 & \textbf{12.91} & \textbf{2.46}\\ 
    \bottomrule
    \end{tabular}
\vspace{-0.6cm}
\end{table}
\indent In this paper, we also focus on capturing longer contexts in contrast to previous work on OOV KWS. This requirement is specially important when considering differently-sized subwords on morphologically-rich language datasets because OOV words are longer than frequent words, as shown in Table \ref{tab:oov_lengths}. Additionally, higher-order \textit{n}-grams can be beneficial for capturing long-term dependencies in an approximated RNNLM. Hence, we introduce an \textit{n}-gram-growing algorithm to our approximation method to facilitate building long-context \textit{n}-gram LM versions of approximated RNNLMs (Section \ref{sec:approx}).\\
\indent In the experiments, we train our baseline LMs and RNNLMs on single- and multi-character subwords. To evaluate these LMs on their OOV detection efficacy, we setup two KWS tasks on Arabic and Finnish datasets using only such OOV keywords that do not appear in the training data (Section \ref{sec:setup} \& \ref{sec:kws}).

%% file: related_work.tex
\section{Related Work}
\subsection{Approximating RNNs to Long-Span \textit{n}-gram LMs}
There exist a few approximation techniques for converting RNNLMs to \textit{n}-gram LMs: variational approximation \cite{Deoras2011}, probability-conversion \cite{Adel2014} and iterative conversion \cite{Adel2014,Arisoy:2014:CNN:2583704.2583722}. Prior work \cite{Adel2014} compared these techniques and the best approximating technique -- iterative conversion -- outperformed other methods using smaller order \textit{n}-grams in a speech recognition task. However, iterative conversion method's effectiveness on low-order word \textit{n}-grams seems sub-optimal for OOVs.  The OOVs are usually longer than frequent words (refer Table \ref{tab:oov_lengths}) and when represented as subwords (like single- and multi-characters) require LMs which can capture long-term information well. Among other methods, the variational approximation method is also not a good fit for our purposes, as it can fail to sample rare subword contexts, which might be present in the subword sequences of OOVs. Hence, we use the probability-conversion method to develop a more efficient method and introduce an \textit{n}-gram-growing variant of the algorithm for approximating subword RNNLMs to larger context sizes.

\subsection{Subword-based Keyword Search for OOV Prediction}
Subword-based Keyword search for OOVs has been performed using either phonetic units like phones, graphones, syllables and sequences of phones \cite{4777893,bulyko2012subword,karakos2014subword,DBLP:journals/taslp/HeBFHJOFP16,DBLP:conf/interspeech/YuS04,DBLP:conf/icassp/SeideYMC04,DBLP:conf/interspeech/LeeT016} or textual single- or multi-character units \cite{DBLP:conf/interspeech/HartmannLMLG14,Khokhlov2017}.\\
\indent The textual approach saves effort spent on generating phonetic representations of OOV words, which has been a common trend across the phonetic approach to KWS systems. Our KWS system is similar to the second category of subword-based KWS systems where we use single- and multi-character units for decoding. Although multi-character subword units have been used for decoding before, single-character units for decoding have been found to obtain poor performance in a KWS system \cite{DBLP:conf/interspeech/HartmannLMLG14}. However, this may be an effect of an ASR system that does not work well on character units. For the datasets considered in this work, we apply tools that allow applying grapheme-based acoustic models for subword ASR improving the performance for character-based ASR compared to word-based ASR \cite{SmitFST}.

%% file: variprob.tex
\section{Approximating RNNLMs to \textit{n}-grams}
\label{sec:approx}

Given a corpus of subword units ($w_i$) and a corresponding vocabulary ($V$), RNNs ($p_r$) can be approximated to \textit{n}-grams ($p_n$) using the probability-conversion (PC) method \cite{Adel2014}. For an \textit{n}-gram history of subwords $h$ and its back-off history $\bar{h}$, PC marginalizes over observed sentence histories ($b$) that precede $h$ in the RNN scoring function $y(w|h)$. $y(w|h)$ is then used in a back-off LM ($p_n^{PC}(w|h)$) to approximate the RNN:
\vspace{-0.2cm}
\begin{align}
\nonumber &y(w|h) = \sum_{b\in H_{*|hw}} \frac{c(bhw)}{c(hw)}\cdot p_r(w|bh)\\
\label{eqn:pc} &p_n^{PC} (w|h) = S\cdot \frac{y(w|h)}{\sum_{v \in V} y(v|h)} + (1 - S)\cdot p_n^{PC}(w|\bar{h})
\end{align}
\noindent Here $H_{*|h}$ is the set of observed sentence histories preceding $h$, $c(x)$ is the count of a sequence $x$, and $0 < S < 1$ is the smoothing factor.\\
\indent As subword LMs for OOVs will need higher-order \textit{n}-grams, the above method can be inefficient due to the normalization term in (\ref{eqn:pc}) and can only produce \text{n}-gram LMs of order three to five effectively. Subverting the normalization calculation, we can also consider the marginalization on sentence histories ($b$) in a novel way for approximating RNNs, whereby,
\begin{align}
&\nonumber p_n^{\textrm{Ours}} (w|h) &=& \sum_{b\in H_{*|h}} p(w,b|h)\\
&\nonumber &=& \sum_{b\in H_{*|h}} p(b|h) p(w|bh)\\
&\label{eqn:ours} &\approx& \sum_{b\in H_{*|h}} \frac{c(bh)}{c(h)}\cdot p_r(w|bh)
\end{align}
\indent The above formula provides a proper distribution with the requirement that RNNLM output vector ($O(bh)$) for every context $bh$ in the corpus is available. For a large corpus ($C$) and vocabulary ($V$), storing and using the complete output vector for every context becomes as resource intensive as (\ref{eqn:pc}), requiring $\mathcal{O}(|C|\cdot|V|)$ storage capacity and operations.\\
\indent To reduce the resource requirements, we only store the probabilities $p_r(w|bh)$ for the context $bh$ under two conditions: the $n$-gram is observed in the corpus ($c(bhw) > 0$), or $w$ is included in $O^{\textrm{top}K}_u(bh)$, the top $K$ probabilities of the RNN's output vector for $bh$ where the next word $u \ne w$. This restriction considers only a part of the output vector instead of the complete output vector, and consequently, lowers the complexity to $\mathcal{O}(|C|\cdot K+1)$. We implement the restriction by considering the contexts $C^{1+\textrm{top}K}$ that fulfill one of the above two conditions: $C^{1+\textrm{top}K} = \{bhw \,|\, c(bhw) > 0 \,\lor\, (\exists u \in V: c(bhu) > 0 \land w \in O^{\textrm{top}K}_u(bh) \land w \ne u)\}$. Hence, (\ref{eqn:ours}) changes to:
\begin{align}
\label{eqn:fast_ours}p_n^{\textrm{Ours}} (w|h)=\sum_{b\in H_{*|h}}\frac{c(bh)}{c(h)}\cdot\mathds{1}_{bhw \in C^{1+\textrm{top}K}}\cdot p_r(w|bh) 
\end{align}
\noindent where $\mathds{1}_{bhw \in C^{1+\textrm{top}K}}$ is the indicator function to represent contexts from the set $C^{1+\textrm{top}K}$. In both (\ref{eqn:ours}) and (\ref{eqn:fast_ours}), calculating $c(bh)$  is not necessary and we can also just calculate the sum of probabilities for an \textit{n}-gram $bhw$.\\
\indent The formulation in (\ref{eqn:fast_ours}) is not a proper distribution, and using small values of $K$ (e.g. $K=0,1$ etc.), the missing probability mass can be high when creating backing-off LMs. Still, for small values of $K$, it provides a further speed up in comparison to (\ref{eqn:ours}) for approximating RNNLMs. \\
\indent Still creating an approximated RNNLM for higher-order \textit{n}-grams using (\ref{eqn:fast_ours}) can be prohibitively expensive. Hence, we embed (\ref{eqn:fast_ours}) in an \textit{n}-gram-growing algorithm \cite{4244538}. This algorithm\footnote{The algorithm (\ref{eqn:fast_ours}) implementation is available at \url{https://github.com/lallubharteja/variKN}} can iteratively grow the \textit{n}-grams in the LM, selecting important \textit{n}-gram contexts using a cost function based on minimum description length of the model and the data. In this algorithm, we can also specify two parameters to limit the number of \textit{n}-grams: the minimum threshold for accepting an \textit{n}-gram, and maximum context size ($n$). For brevity, we do not describe details of the growing algorithm here, but refer readers to \cite{4244538}.

%% file: experimental_setup.tex
\section{Experimental Setup}
\begin{table*}[!ht]
  \centering
  \small
  \caption{\label{tab:MTWV} KWS performance for both morphs and single characters is presented on Arabic and Finnish datasets along with language model size in number of \textit{n}-grams. The table reports MTWV and Lattice Recall calculated on first-pass decoded lattices for KNV, RNN5, RNNV and its linear interpolation (KNV+RNNV) with equal weights.}
  \vspace{-0.3cm}
  \begin{tabular}{l l c c c c c c}
    \hline
    \multirow{2}{*}{Segmentation} & \multirow{2}{*}{LM} & \multicolumn{3}{ c }{Arabic} & \multicolumn{3}{c}{Finnish} \\
    \cline{3-8}
     & & MTWV & Lattice Recall & \#N-grams & MTWV & Lattice Recall & \#N-grams\\
    \hline\hline
    \multirow{5}{*}{morphs} & KNV & 0.551 & 0.505 & 6.13M & 0.317 & 0.299 & 4.18M\\
    & RNN5 (PC) & 0.245 & 0.244 & 279K & 0.128 & 0.116 & 1.24M\\
    & KNV+RNN5 (PC)& 0.560 & 0.531 & 6.27M & 0.308 & 0.289 & 4.85M\\
    \cline{2-8}
    & RNNV (Ours) & 0.328 & 0.350 & 109K & 0.202 & 0.183 & 1.46M\\
    & KNV+RNNV (Ours)& \textbf{0.591} & \textbf{0.564} & 6.19M & \textbf{0.321} & \textbf{0.303} & 4.71M\\
    \hline
    \hline
    \multirow{5}{*}{characters} & KNV & 0.561 & 0.551 & 5.56M & 0.686 & 0.650 & 6.75M\\
    & RNN5 (PC)& 0.595 & 0.610 & 501K & 0.704 & 0.693 & 636K\\
    & KNV+RNN5 (PC)& 0.646 & 0.634 & 5.87M & 0.715 & 0.696 & 7.04M\\
    \cline{2-8}
    & RNNV (Ours)& 0.659 & \textbf{0.689} & 618K & 0.699 & 0.687 & 548K\\
    & KNV+RNNV (Ours)& \textbf{0.692} & 0.673 & 5.90M & \textbf{0.738} & \textbf{0.721}& 6.94M\\
    \bottomrule
    \vspace{-0.8cm}
    \end{tabular}
\end{table*}
\label{sec:setup}
We setup the experiments on publicly-available datasets from two languages: Arabic and Finnish. Both languages are morphologically rich, leading to quite a few OOV words in the datasets (OOV rate $\sim$ 2\%).
\subsection{Datasets}
\label{ssec:data}
For Arabic acoustic models, we used the training corpus from the MGB-2 challenge \cite{Ali0GMMRZ16}, consisting of 1,200 hours of Aljazeera's television programs data. For testing, we used the MGB-2 development set, which has eight hours of data and 57k words. For Arabic language models, we used a corpus of 130 million tokens obtained from the Aljazeera's website. This text contains around 1.4 million unique words. For Finnish acoustic models, we used 1500 hours of Finnish audio data from three different data sets, namely, the Speecon corpus \cite{conf/lrec/IskraGMHDK02}, the Speechdat database \cite{speechdat} and the parliament corpus \cite{parl}. For testing, we used a set of broadcast news from the Finnish national broadcaster (Yle) containing 5 hours of speech and 35k words \cite{parl}. For Finnish, we train the language models on the Finnish Text Collection \cite{FTC}. The training set consists of 143M tokens with 4.2M unique types.
\subsection{Keyword Search Setup}
The subword KWS requires a subword-based ASR system. The ASR system is set up using the Kaldi toolkit \cite{Povey_ASRU2011} in a similar fashion to the subword systems presented in \cite{Smit2017}.  These systems apply grapheme-based acoustic models that can generate pronunciations for all subwords. This setup also requires modification of the weighted finite-state transducer of the lexicon (L-FST) for treating subword units similarly to words in Kaldi. For the details of the modification, see \cite{SmitFST}.\\
\indent For setting up the subword keyword search task, we create a list of keywords from the language-specific test sets. We extract the OOV words from the evaluation set, but remove any OOVs which have only a single character or have a character that is not present in the words of the training set; and for Finnish are spelled incorrectly. Thus, we obtain 449 and 661 OOV keywords for the Arabic and Finnish test sets, respectively.\footnote{We publish the OOV lists for these tasks at \url{https://github.com/lallubharteja/KWS-Scripts}}\\
\indent Next, we segment these keywords to single- and multi-character subwords; and apply scripts from Kaldi's openKWS system \cite{Trmal2017} to setup the KWS task for the different segmentations. To evaluate our models in keyword search, we use the Maximum Term Weighted Value (MTWV) on the first-pass lattices. MTWV is a popular KWS metric which incorporates both keyword-specific misses and false alarms. For further details on MTWV, see \cite{6707728}. We also measure each lattice's keyword recall, which measures the amount of correctly retrieved keywords out of the relevant ones.
\input{lms.tex}

%% file: lms.tex
\subsection{Building Language Models}
\label{ssec:lms}
As subwords, we use single \textbf{characters} and the multi-character units created by Morfessor Baseline \cite{Creutz2002,morf2013}. For brevity, we refer to the latter units as \textbf{morphs}, although they do not correspond to linguistic morphemes. For Arabic, we mark the right end of subwords except when at the end of a word. E.g. (international = inter+ nation+ al). For Finnish, we mark both the left and right ends of subwords except when at the beginning or the end of a word. E.g. (international = inter+ +nation+ +al). These choices are based on prior work \cite{Smit2017}, which shows that the respective marking schemes for Arabic and Finnish datasets outperform other schemes.\\
\indent As the \textit{n}-gram LM baselines, we train the Kneser-Ney \cite{479394} smoothed variable-length \textit{n}-grams (\textbf{KNV}) using the VariKN toolkit \cite{4244538}. For first-pass decoding, the RNNLMs, built similarly to the \textit{small} architectures from \cite{Smit2017}, are approximated using RNN approximation method from Section \ref{sec:approx} with $K=3$ and denoted as \textbf{RNNV}. For RNNV LMs, we use a threshold of $0.1$ for both types of subword segmentations. For Arabic, the best \textbf{RNNV} is obtained for $n$'s 13 and 6 for morphs and characters respectively. For the Finnish, $n$'s 11 and 5 prove best for morphs and character \textbf{RNNV} models, respectively.\\
\indent For comparison with our method, we also build LMs using the Probability-Conversion method (PC). Creating higher-order LMs can be expensive with the PC method, so we approximate RNNLMs to 5-gram models (\textbf{RNN5}) using this method. The \textbf{RNN5} are constrained to have similar number of 5-grams as in the corresponding \textbf{RNNV} to keep the model strength of 5-gram LMs comparable. On the Arabic and Finnish text, creating \textbf{RNN5} is 2.3/12 and 1.6/6.7 times slower than building the corresponding \textbf{RNNV} with characters/morphs respectively.\\
\indent Additionally, we linearly interpolate \textbf{RNNV} and \textbf{RNN5} LMs with KNV using equal weights, creating \textbf{KNV+RNNV} and \textbf{KNV+RNN5}. We note that the RNNLM approximations and their interpolated versions applied to the first-pass had a worse ASR performance than applying \textbf{KNV} for the same, but are not presented here for conciseness. The perplexities of these models cannot be compared, as the approximated RNNs are not proper distributions.

%% file: kws_exp.tex
\begin{figure*}[!ht]
  \begin{subfigure}{0.5\textwidth}
  \centering
    \includegraphics[height=4.2cm,trim={0.2cm 0cm 0.2cm 0},clip]{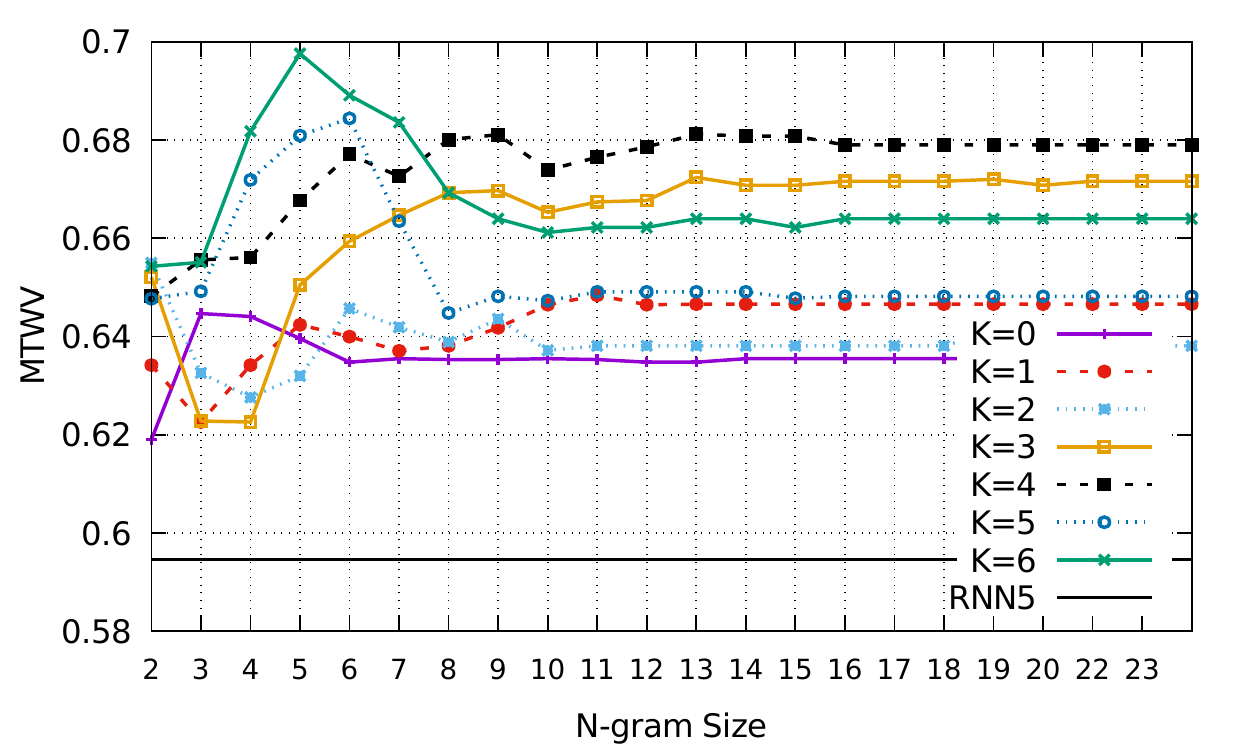}
    \subcaption{Arabic}
  \vspace{-0.4cm}
  \end{subfigure}%
  \hfill
  \begin{subfigure}{0.5\textwidth}
  \centering
    \includegraphics[height=4.2cm,trim={0.2cm 0cm 0.2cm 0},clip]{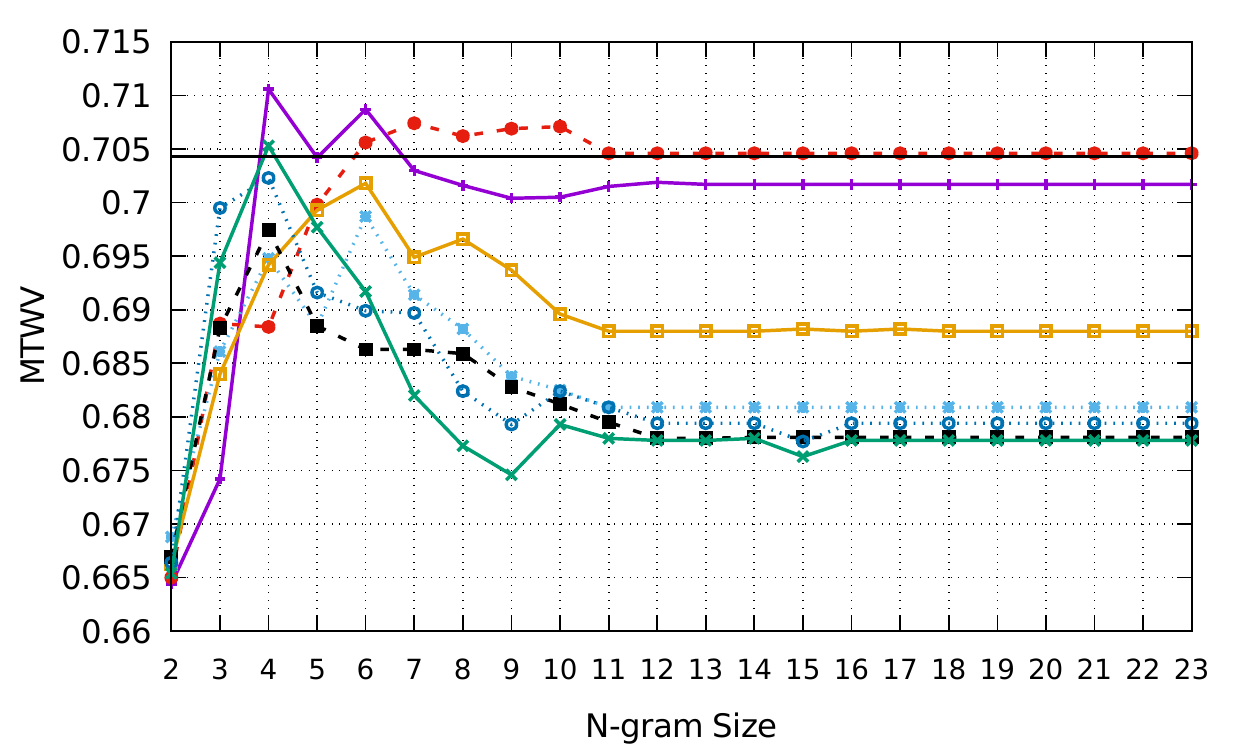}
    \subcaption{Finnish}
  \vspace{-0.4cm}
  \end{subfigure}
    \caption{\label{fig:topk}Single-character \textbf{RNNV}'s MTWV measured by varying the $K$ values in the top-$K$ step of the algorithm (\ref{eqn:fast_ours}) for Arabic and Finnish KWS tasks. \textbf{RNN5} as constructed in Section \ref{ssec:lms} is shown by the thick horizontal line.}
\vspace{-0.7cm}
\end{figure*}
\begin{figure}[t]
  \centering
    \includegraphics[width=\linewidth,height=4.5cm,trim={0.1cm 0.4cm 0.1cm 0},clip]{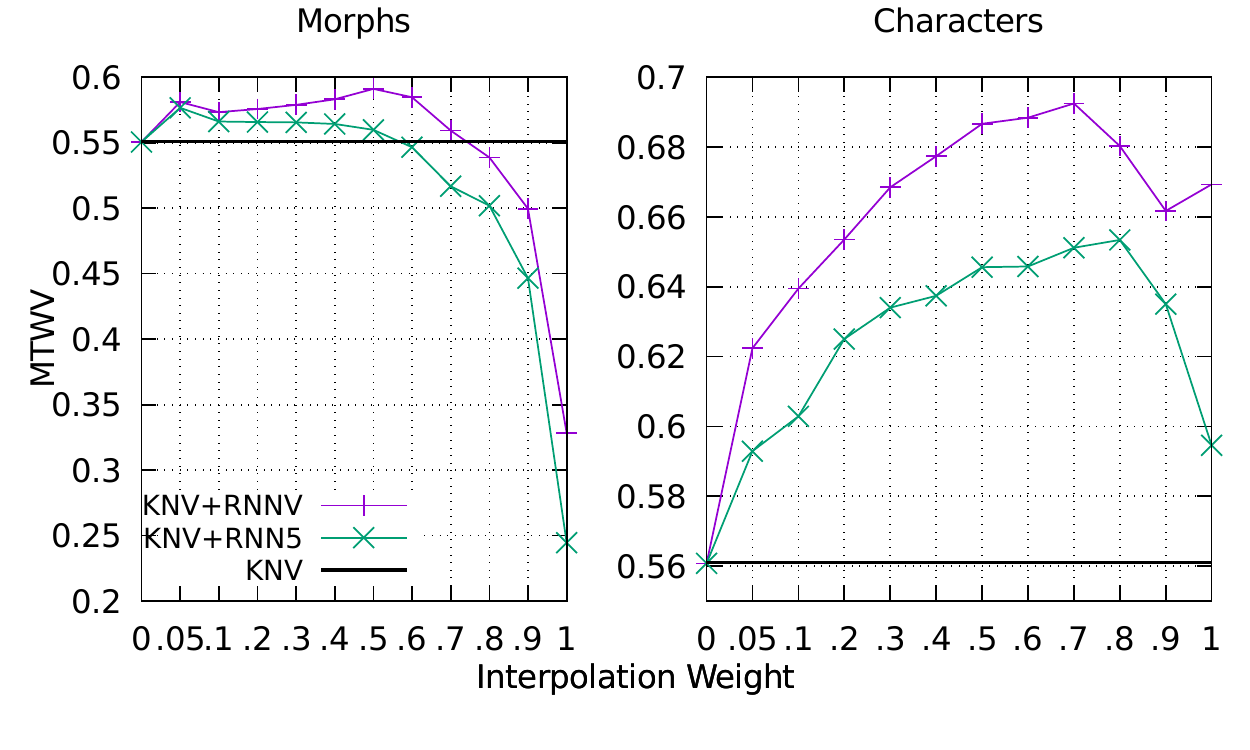}
    \vspace{-0.4cm}
    \caption{Interpolated LMs' MTWV measured by varying the interpolation weight of RNNLMs approximations in in \textbf{KNV+RNN5} and \textbf{KNV+RNNV} models for Arabic KWS task. \textbf{KNV} MTWV is displayed for reference.}
    \label{fig:weights}
\vspace{-0.6cm}
\end{figure}
\section{Keyword Search Experiments}
\label{sec:kws}
In this work, we perform KWS experiments for Arabic and Finnish OOV words from the evaluation set, which are the hardest to predict as they are unseen in the training set. The word-based LMs that are devoid of any phonetic information will fail at predicting them and have a zero MTWV. Hence, we setup a subword-based keyword search on the above datasets.\\
\indent Table \ref{tab:MTWV} compares the different LMs from Section \ref{ssec:lms} on the Arabic and Finnish OOV Keyword Search. These LMs are built using morph and single-character subwords. The results in Table \ref{tab:MTWV} show that single-character models perform better than the morph-based models mostly because OOVs can be better represented at character level than with morphs. Similarly, the RNNLM approximations (\textbf{RNN5} and \textbf{RNNV}) have a more competitive performance with \textbf{KNV} at character-level than with the morph units. Furthermore, both the interpolated models (\textbf{KNV+RNN5} and \textbf{KNV+RNNV}) outperform \textbf{KNV} on characters with larger improvements (at least 15.1\% on Arabic and 4.2\% on Finnish) than on morphs, where the improvments at best can be 7.2\% on Arabic and 1.2\% on Finnish.\\
\indent On Arabic and Finnish KWS, \textbf{KNV+RNNV} achieves larger improvements over \textbf{KNV} for different subwords on the Arabic KWS (7.2\%--23.3\%) than on the Finnish KWS (1.2\%--7.6\%). Improved \textbf{KNV} performance on Finnish might be dependent on the quality of the underlying acoustic models. In particular, the Finnish ASR has a larger and cleaner (Speecon and Speechdat transcripts are verified) dataset than used in the Arabic ASR. Also, Finnish has a simpler phonetic structure than Arabic. \\
\indent Across the different subword units, \textbf{RNNV} mostly performs better than \textbf{RNN5} on MTWV, except when using characters on the Finnish KWS. The performance differences are larger across Arabic and Finnish when comparing the interpolated models, with \textbf{KNV+RNNV} performing the overall best. This shows the benefits of using the proposed method, which has a different scoring scheme and access to higher-order \textit{n}-gram contexts than the probability-conversion method.
\section{Analysis}
In this section, we analyse KWS performance's sensitivity with respect to important parameters involved in construction of \textbf{RNNV} and the interpolated models. For \textbf{RNNV}, we look at the choice of $K$ in the top-$K$ step of the algorithm (\ref{eqn:fast_ours}) and the \textit{n}-gram context size. For the interpolated models, we consider the interpolation weight of the RNNLM approximations in \textbf{KNV+RNN5} and \textbf{KNV+RNNV} models.
\subsection{\textit{N}-gram size \& top-\textit{K} values in RNNLM approximation}
In Figure \ref{fig:topk}, we report KWS performance when varying the $K$ from 1 to 6 and $n$ from 5 to 23 for the character models while keeping the growing algorithm's threshold fixed at 0.1. A fixed threshold forces an (\textit{n+1})-gram LM to have the same n-grams as the \textit{n}-gram LM had before growing to the (\textit{n+1})-grams.\\
\indent On both Arabic and Finnish, we find \textbf{RNNV} for $K=0$ can outperform \textbf{RNN5}. Lowering the $K$ allows faster construction of \textbf{RNNV} and thus, improves benefits over PC-based method. Overall, the performance for all $K$s seem to vary more with $n<4$ and then stabilize with $n>=4$ for some $K$s even outperforming the smaller $n$. These improvements suggest that the system can benefit from longer contexts $n>=4$. These observations suggest that choosing $K$ and $n$ are important to enable efficient and improved performance of \textbf{RNNV}.\\
\subsection{Varying the RNNLM Interpolation Weight}
Figure \ref{fig:weights} show the variation of MTWV against different interpolation weights (in the range of $[0,1]$) of \textbf{RNN5} and \textbf{RNNV} in their respective interpolated models for the Arabic KWS task. For some interpolation weights, \textbf{KNV+RNN5} and \textbf{KNV+RNNV} performance can be further improved than results observed in Table \ref{tab:MTWV}. In most cases, \textbf{KNV+RNNV} achieves a better performance than \textbf{KNV+RNN5} suggesting that \textbf{RNNV} is able to complement \textbf{KNV} better than \textbf{RNN5}. Similar trends were also observed for the Finnish KWS task.

%% file: conclusion.tex
\section{Concluding Remarks}
In this paper, we introduced a new efficient technique inspired from probability-conversion method to approximate RNNLMs to \textit{n}-gram LMs. We also extended this technique using \textit{n}-gram-growing algorithm to better handle OOVs and create better long-span approximations of RNNLM.\\
\indent Using multi- and single-character subwords, we constructed interpolated LMs using conventional \textit{n}-gram and approximated RNN models. We applied these models on first-pass-based Arabic and Finnish Keyword Search for OOVs, which are the hardest to predict. We observed that our method, which had longer contexts, complemented the conventional \textit{n}-gram LMs better than the probability-conversion method. In addition, single-character-based LMs outperformed the morph-based LMs and using the proposed method single-character models performed the best overall. In future, we would also like to investigate the effect of rescoring on this KWS setup.\\
\indent While predicting OOVs in a high-resource scenario, we were able to achieve MTWVs higher than IARPA's BABEL Program aim of 0.5 MTWV. Still, we want to explore if similar performance can be obtained on an under-resourced scenario as prescribed by IARPA's BABEL program.